\documentclass[letterpaper, 10 pt, conference]{ieeeconf}  

\IEEEoverridecommandlockouts                              

\usepackage{graphics} 
\usepackage{graphicx}
\usepackage{makecell}
\usepackage{amsmath}
\usepackage{amssymb}
\usepackage{siunitx}
\usepackage[nolist]{acronym}

\usepackage{pifont}  
\newcommand{\cmark}{\ding{51}}  
\newcommand{\xmark}{\ding{55}}  

\title{\LARGE \bf
LiHRA: A LiDAR-Based HRI Dataset for Automated Risk Monitoring Methods
}

\author{Frederik Plahl$^{1,2*}$, Georgios Katranis$^{2}$, Ilshat Mamaev$^{1}$, Andrey Morozov$^{2}$
\thanks{$^{*}$ Corresponding author}
\thanks{$^{1}$Proximity Robotics \& Automation GmbH, 
    76327 Pfinztal,
    Germany
    {\tt\small \{last\}@proximityrobotics.com}}%
\thanks{$^{2}$Institute of Industrial Automation and Software Engineering,
    University of Stuttgart,
    70550 Stuttgart,
    Germany
    {\tt\small \{first.last\}@ias.uni-stuttgart.de}}%
\thanks{Project page:\newline{\tt\small https://proximityrobotics.github.io/LiHRA}}
\thanks{This paper presents scientific results from the research project "CogniSafe3D" (E!6085), conducted within the framework of the European funding program Eurostars. The German partners are funded by the Federal Ministry of Research, Technology and Space under grant number 01QE2426A.}
\thanks{© 2025 IEEE. This is the author’s accepted manuscript of a paper that will appear in the Proceedings of the IEEE/RSJ International Conference on Intelligent Robots and Systems (IROS 2025). The final published version is available via IEEE Xplore: https://ieeexplore.ieee.org/}
}

\begin{document}
\begin{acronym}
\end{acronym}

\maketitle
\thispagestyle{empty}
\pagestyle{empty}

\begin{abstract}
We present LiHRA, a novel dataset designed to facilitate the development of automated, learning-based, or classical risk monitoring (RM) methods for Human-Robot Interaction (HRI) scenarios. The growing prevalence of collaborative robots in industrial environments has increased the need for reliable safety systems. However, the lack of high-quality datasets that capture realistic human-robot interactions, including potentially dangerous events, slows development. LiHRA addresses this challenge by providing a comprehensive, multi-modal dataset combining 3D LiDAR point clouds, human body keypoints, and robot joint states, capturing the complete spatial and dynamic context of human-robot collaboration.
This combination of modalities allows for precise tracking of human movement, robot actions, and environmental conditions, enabling accurate RM during collaborative tasks.
The LiHRA dataset covers six representative HRI scenarios involving collaborative and coexistent tasks, object handovers, and surface polishing, with safe and hazardous versions of each scenario. In total, the data set includes 4,431 labeled point clouds recorded at 10 Hz, providing a rich resource for training and benchmarking classical and AI-driven RM algorithms.
Finally, to demonstrate LiHRA's utility, we introduce an RM method that quantifies the risk level in each scenario over time. This method leverages contextual information, including robot states and the dynamic model of the robot.
With its combination of high-resolution LiDAR data, precise human tracking, robot state data, and realistic collision events, LiHRA offers an essential foundation for future research into real-time RM and adaptive safety strategies in human-robot workspaces.

\end{abstract}

\section{Introduction}
\label{sec:Introduction}
Human-Robot Interaction (HRI) is becoming increasingly prevalent in industrial settings. However, ensuring the safety of human workers remains a critical challenge. International standards, such as ISO 10218-2:2011 and ISO/TS 15066:2016, emphasize Risk Assessment (RA) as the first step in identifying and mitigating hazards in collaborative environments. The recommended practices rely on manual and often subjective evaluations conducted by domain experts \cite{ISO15066, ISO10218}. Researchers argue that the methods currently accepted by these standards are difficult to apply to HRI due to the field’s inherent complexity, limited experience, challenges in predicting human behavior, and the intricacies of estimating collision criticality \cite{huck_risk_2021}. Furthermore, they highlight the need to incorporate human factors into hazard analysis and RA for human-robot collaboration (HRC) as a more human-centered design is essential for guaranteeing safety \cite{giallanza_occupational_2024}. In the process of RA, different methods for risk mitigation, monitoring the safety of the system, can be suggested by the experts. However, the development of these Risk Monitoring (RM) methods demands the availability of datasets that provide similar data to what would be available during the operation of HRI in the industry.

 \begin{figure}[t]
    \centering
        \includegraphics[width=0.9\columnwidth]{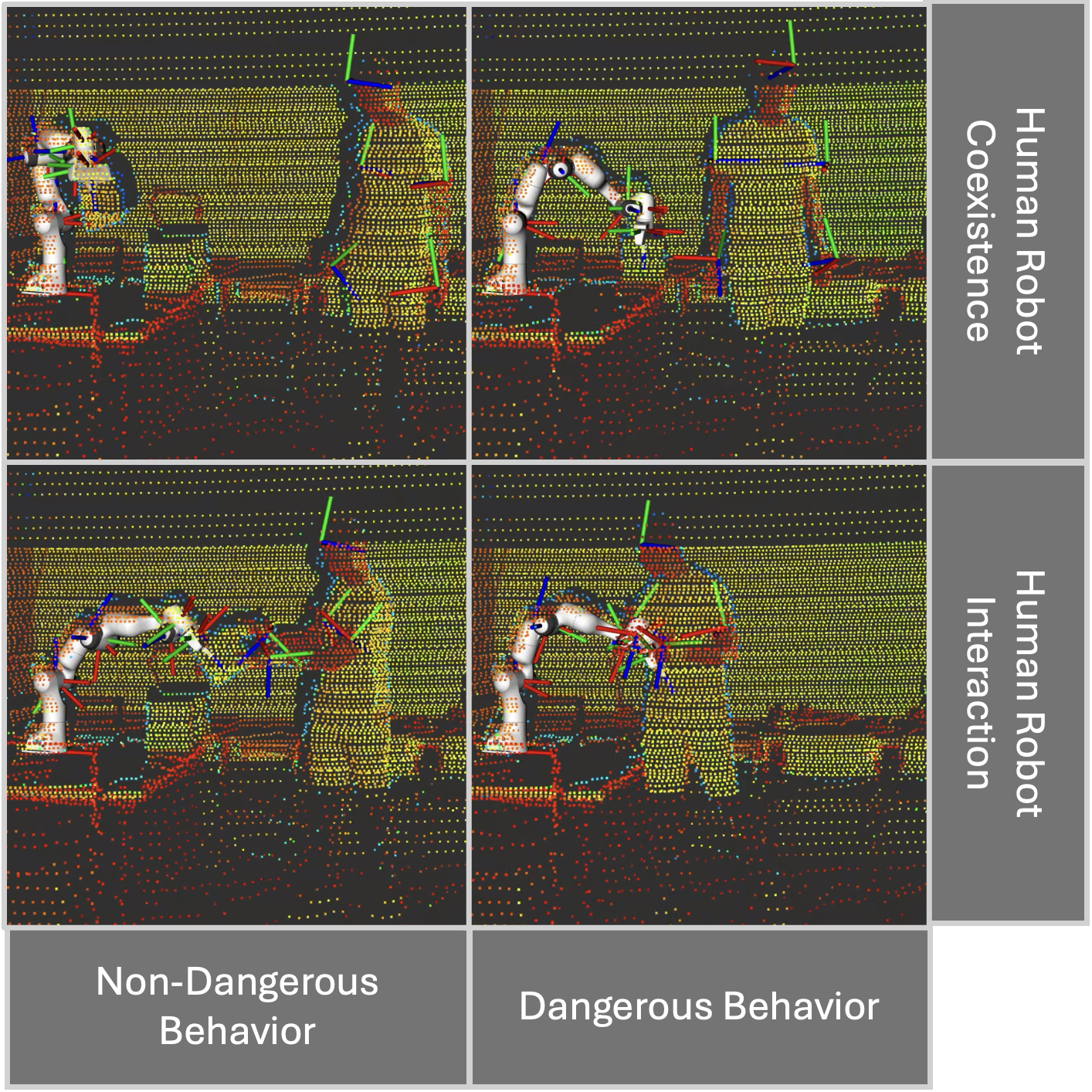} 
    \caption{LiHRA includes labeled point clouds from both safe and hazardous scenarios. The top row depicts a coexisting scenario: a human safely observing the robot (left) and an inattentive human colliding with it (right). The bottom row shows a handover scenario: a safe item transfer (left) and the robot approaching an unaware human (right).}
    \label{fig:dataset-picture}
\end{figure}

Among the classical methods, such as FMEA and HAZOP, simulation-based and Computer-Aided Risk Assessment (CARA) tools have been developed to assist experts in conducting RA while accounting for human behavior \cite{huck_testing_2022, awad_integrated_2017, araiza-illan_systematic_2016}. However, these approaches often lack realism due to simplified models and constraints in sensor integration \cite{alenjareghi_safe_2024}. Additionally, the complexity of human behavior makes simulating an entire process practically infeasible \cite{Huck_Testing_System_Safety}. To address these limitations, researchers are exploring AI-driven RA techniques \cite{terra_safety_2020}. Nonetheless, the application of AI to RA remains underexplored \cite{alenjareghi_safe_2024}. The effectiveness of AI-based RA methods heavily depends on high-quality datasets, as the accuracy of AI-driven tools is directly influenced by the data they are trained on. If the data is biased, incomplete, or outdated, the resulting RA may be unreliable \cite{alenjareghi_safe_2024}. The required data for the development of more advanced algorithms for RA in HRI, like robot joint positions, velocities, and torque data, as well as human limb positions, is captured by our dataset. This bridges the gap and enables the development of more advanced learning based algorithms.

For the vision-based component of our dataset, we selected a 3D LiDAR sensor. A key advantage of 3D LiDAR for RA is its reliability in high-risk applications, a benefit demonstrated not only in autonomous driving \cite{zhao_fusion_2020, zermas_fast_2017, li_lidar_2020}, but also in HRI \cite{arora_using_2024-1, podgorelec_lidar-based_2023}. Given the similarities in object sizes, scene dynamics, and environmental constraints between autonomous driving and HRI, we argue that insights from the former can be effectively applied to the latter. Additionally, 3D LiDAR offers the advantage of GDPR compliance, as it does not capture personally identifiable visual data. However, our primary motivation for integrating it into this dataset lies in its technical strengths and its feasibility for certification in industrial safety applications. By leveraging 3D LiDAR, our dataset provides robust perception capabilities, complementing other sensory modalities to enhance RA in HRI.

To conduct relevant RA for HRI, accurate information about multiple factors, like human behavior, technological interactions and the conditions of the environment (e.g., state of the robot, distance between human and robot) are needed \cite{alenjareghi_safe_2024}. Most of these factors can only be assessed by monitoring the state of the robot (i.e., position, velocity, effort of the joints), the position of the human body including limbs, and a general view on the environment of the scene.
Despite the growing popularity of HRI, no existing datasets, to the best of our knowledge, meet the specific requirements for HRI RA and especially on-line RM. To bridge this gap, we introduce \textbf{LiHRA}, a novel LiDAR-based dataset designed explicitly to enable the development of automated RA in HRI. LiHRA provides precise, well-labeled data that captures HRI dynamics, enabling researchers and practitioners to develop, test, and refine next-generation RA and RM methodologies.

The LiHRA dataset consists of:

\begin{itemize}
    \item 3D LiDAR point clouds, capturing the HRI scene using an image-grade \textit{Seyond Falcon Kinetic LiDAR},
    \item human keypoints (head, hands and shoulders), recorded with \textit{HTC VIVE Tracker 3.0},
    \item robot joint states, velocities, and efforts, extracted from a \textit{Franka Emika Robot (FER)}.
\end{itemize}

The dataset includes safety-critical HRI scenarios such as object handovers, collaborative tasks, and shared workspace interactions. LiHRA also includes intentional contact (e.g., physical contact inherent in handovers) and unintentional collisions, which are staged to simulate potentially dangerous interactions between the human and the robot.

To demonstrate the dataset’s applicability for automated RM, we propose a method to quantify the risk level in each provided HRI scenario as a function over time. This method incorporates the contextual information, including human keypoints, robot joint positions, velocities, and torques, as well as derived data such as distances and end-effector velocities. Hazard indicators are defined using heuristic functions that process these inputs and estimate possible risks. To ensure compliance with standards, the limit values for these functions are aligned with ISO/TS 15066:2016. Furthermore, the collision severity in case of collaboration is estimated, a measure of the severity of the risk. 

The main contributions of this paper can be summarized in the following three points:
\begin{itemize}
    \item To the best of our knowledge, we present the first dataset that integrates LiDAR point clouds, joint states from a cobot arm, and human keypoints captured using an \textit{HTC VIVE Tracker 3.0}. This dataset lays the foundation for developing both classical and AI-driven automated RM methods, particularly by demonstrating a framework for RM as a key aspect of risk mitigation.
    \item We propose a new automated RM method, based on formulas derived from applicable safety standards. This shows the usability and also provides risk indices which can be used in combination with the dataset to train learning based methods.
\end{itemize}

The remainder of this paper is structured as follows: Section \ref{sec:Related-Work} reviews related work, Section \ref{sec:The-LiHRA-Dataset} provides a detailed description of LiHRA, Section \ref{sec:Experimental-Evaluation} presents the RM method, and Section \ref{sec:Conclusions} concludes the paper.

\section{Related Work}
\label{sec:Related-Work}
\begin{table*}[t]
\caption{Comparison of HRI datasets.}
\label{tab:related_work}
\begin{center}
\begin{tabular}{c|c c c c c c c c}
\hline
Dataset & \makecell{LiDAR} & \makecell{Robot\\ Joint\\States} & \makecell{Human\\ Skeleton} & \makecell{Human\\Joints} & \makecell{Human\\viewing\\angle} & \makecell{MoCap\\ System} & \makecell{Collisions /\\ Intended\\Contact} & \makecell{Risk\\calculation\\ method}\\
\hline
COVERED \cite{munasinghe_covered_2022}  & \cmark & \xmark & \xmark & \xmark     & \xmark   & \xmark                             & \xmark    & \xmark\\
PIA-NNs \cite{lin_prior_2024}      & \cmark & \cmark & \xmark & \xmark     & \xmark   & \xmark                             & \xmark    & \xmark\\
InHARD \cite{dallel_inhard_2020}    & \xmark & \xmark & \cmark & 17         & \xmark   & \makecell{Perception\\Neuron}     & \xmark    & \xmark\\
INTERACT \cite{yasar_posetron_2024}& \xmark & \cmark & \cmark & unknown    & (\cmark) & OptiTrack                          & \xmark    & \xmark\\
FACT HRC \cite{tian_crafting_2023}    & \xmark & \cmark & \cmark & 25         & \xmark   & \xmark                          & (\cmark)  & \xmark\\
HARMONIC \cite{newman_harmonic_2022}& \xmark & \cmark & \cmark & 18+42      & \cmark   & \xmark                            & \xmark    & \xmark\\
CHICO \cite{sampieri_pose_2022}      & \xmark & \xmark & \cmark & 15         & \xmark   & \xmark                           & \cmark    & \xmark\\
HARPER \cite{avogaro_exploring_2024}    & \xmark & \cmark & \cmark & 21         & \xmark   & OptiTrack                     & \cmark    & \xmark\\
\hline
\textbf{LiHRA (ours)}   & \cmark & \cmark & \cmark & 5         & \cmark   & \makecell{HTC VIVE\\Tracker 3.0}  & \cmark    & \cmark\\
\hline
\end{tabular}
\end{center}
\end{table*}

While there appears to be no dataset available, focusing on the RM in HRI, we need to take a look on HRI datasets with focus on different domains and discuss their applicability to our use case. The requirements for the dataset are based on how data would possibly be captured in an industrial HRI setting and what data is relevant for an RM method to estimate the risk. As a capturing method, LiDARs have shown to be applicable (as argued in Section \ref{sec:Introduction}). Representations of relevant involved actors in an HRI scenario can be derived from human keypoints and robot joint states.
In Table \ref{tab:related_work}, a comparison between LiHRA and existing datasets of the similar domain is shown and evaluated based on the set requirements.

Most existing HRI datasets rely on RGB and RGB-D cameras, with the exceptions of COVERED \cite{munasinghe_covered_2022} and PIA-NNs \cite{lin_prior_2024}, which utilize LiDAR-based segmentation. COVERED focuses on real-time human-robot segmentation in LiDAR point clouds, while PIA-NNs employs a segmentation algorithm that is leveraged through simulated point clouds from a digital replication of the scene. However, both datasets suffer from limited scene understanding compared to LiHRA, and they do not capture human limb positions, robot joint states, and viewing angles, making them less suitable for automated RA development.

Other HRI datasets, such as InHARD \cite{dallel_inhard_2020}, provide detailed human keypoints using a Motion Capture (MoCap) system but lack robot joint state data. Conversely, datasets like INTERACT \cite{yasar_posetron_2024}, a dataset made for human motion prediction and FACT HRC \cite{tian_crafting_2023} do include robot motion information, with FACT HRC focusing on social cues in HRI using a teleoperated mobile robot. While these datasets contain the necessary information for RA, they lack hazardous scenarios. FACT HRC records minor, unintentional collisions, but these are too rare and insignificant to support the development of new automated RA methods.

The HARMONIC \cite{newman_harmonic_2022} dataset introduces teleoperated assistive eating tasks with a cobot manipulating a fork, presenting an inherently risky scenario. However, this is not representative of typical industrial HRI settings, and no intentional or unintentional collisions are explicitly documented.

To our knowledge, CHICO \cite{sampieri_pose_2022} and HARPER \cite{avogaro_exploring_2024} are the only datasets containing both intentional and unintentional collisions. Intentional collisions occur naturally in contact-based tasks (e.g., handovers), whereas unintentional collisions happen accidentally, such as when a robot’s object-picking movement leads to a contact with a nearby standing human. These unintended collisions are particularly relevant to RA due to their higher impact potential at greater robot speeds. However, CHICO lacks robot joint state data, making it unsuitable for our goals. HARPER solves this issue by tracking the robot with a MoCap system, which it uses for their goal to incorporate collision prediction from the robot’s perspective. However, it does not include 3D LiDAR, a crucial aspect that LiHRA provides, making our dataset a more comprehensive solution for advancing automated RA methods.

Among the existing datasets, LiHRA appears to be the only one that fulfills all the essential requirements for developing automated RM methods, whether through AI-driven approaches or mathematical models directly derived from safety standards.

\section{The LiHRA Dataset}
\label{sec:The-LiHRA-Dataset}

The focus of the LiHRA dataset is enabling the development of automated RM methods. To achieve this, we used an image grade 3D LiDAR \textit{Seyond Falcon Kinetic}, a cobot \textit{Franka Emika Robot (FER)} and a MoCap system build with \textit{HTC VIVE Tracker 3.0}. In Section \ref{sec:Hardware}, the setup and justification for the selection of this specific hardware are presented. 

Using this hardware, a recording pipeline has been constructed, including components for automatic external calibration of the cobot and the MoCap trackers with respect to the LiDAR. A detailed overview about this pipeline is provided in Section \ref{sec:Pipeline}.

The dataset consists of multiple different scenarios which are closely related to actions that can happen in an industrial context. Section \ref{sec:Scenarios} gives a detailed overview on the scenarios recorded in the dataset. In total, the dataset consists of 4,431 labeled point clouds recorded at around 10 FPS.

\subsection{Hardware}
\label{sec:Hardware}
LiDARs are highly effective for precise obstacle detection, environmental mapping and human detection \cite{bilik_comparative_2023, li_lidar_2020}. Resulting in widely spread appliance in the also safety critical autonomous driving field. For this, LiDARs need to meet certain requirements \cite{dai_requirements_2022} which can be transferred to the HRI safety. Consequently, as mentioned in the introduction of this paper (Section \ref{sec:Introduction}), 3D LiDARs are also being used in the safety critical field of HRI \cite{arora_using_2024-1, podgorelec_lidar-based_2023}. In alignment with this application, we have opted to utilize the LiDAR system \textit{Seyond Falcon Kinetic}, which is currently primarily used in the automotive industry.

Assessing the risk in an HRI scenario requires the knowledge of the human's position. One can assume, that future RM solutions depend on a pipeline as i.e., collecting LiDAR point clouds, estimate human keypoints based on the point clouds collect the joint states the robot provides and estimate a risk based on that. If we used an existing Human-Pose Estimation (HPE) method to label the keypoints of the human, logically an AI model trained on this dataset can not provide more precise human keypoints than an HPE method we would use for labeling the data. To get the most accurate data for ground truth data, we decided to use a MoCap system build with trackers used in virtual reality games (\textit{HTC VIVE Tracker 3.0, Lighthouse 2.0}) to provide human keypoints. The accuracy of these trackers lie in the sub centimeter range \cite{niehorster_accuracy_2017}, making them precise enough for the task.

The robot in HRI scenarios is mostly a cobot, which can be used to interact safely close to humans, due to an inherently safe design and sensing capabilities \cite{guertler_when_2023}. To address these requirements, we chose to use the \textit{Franka Emika Robot (FER)} since it not only falls under the definition of a cobot, but also provides detailed information about its dynamics, which can be helpful in RM and is widely used in the research community. As a controller for the cobot, a cartesian impedance controller is used. Controllers like this are best suitable for interactions with humans since they are to some extend compliant to external forces from the human.

\begin{figure}[thpb]
    \centering
    \framebox{
        \includegraphics[width=0.9\columnwidth]{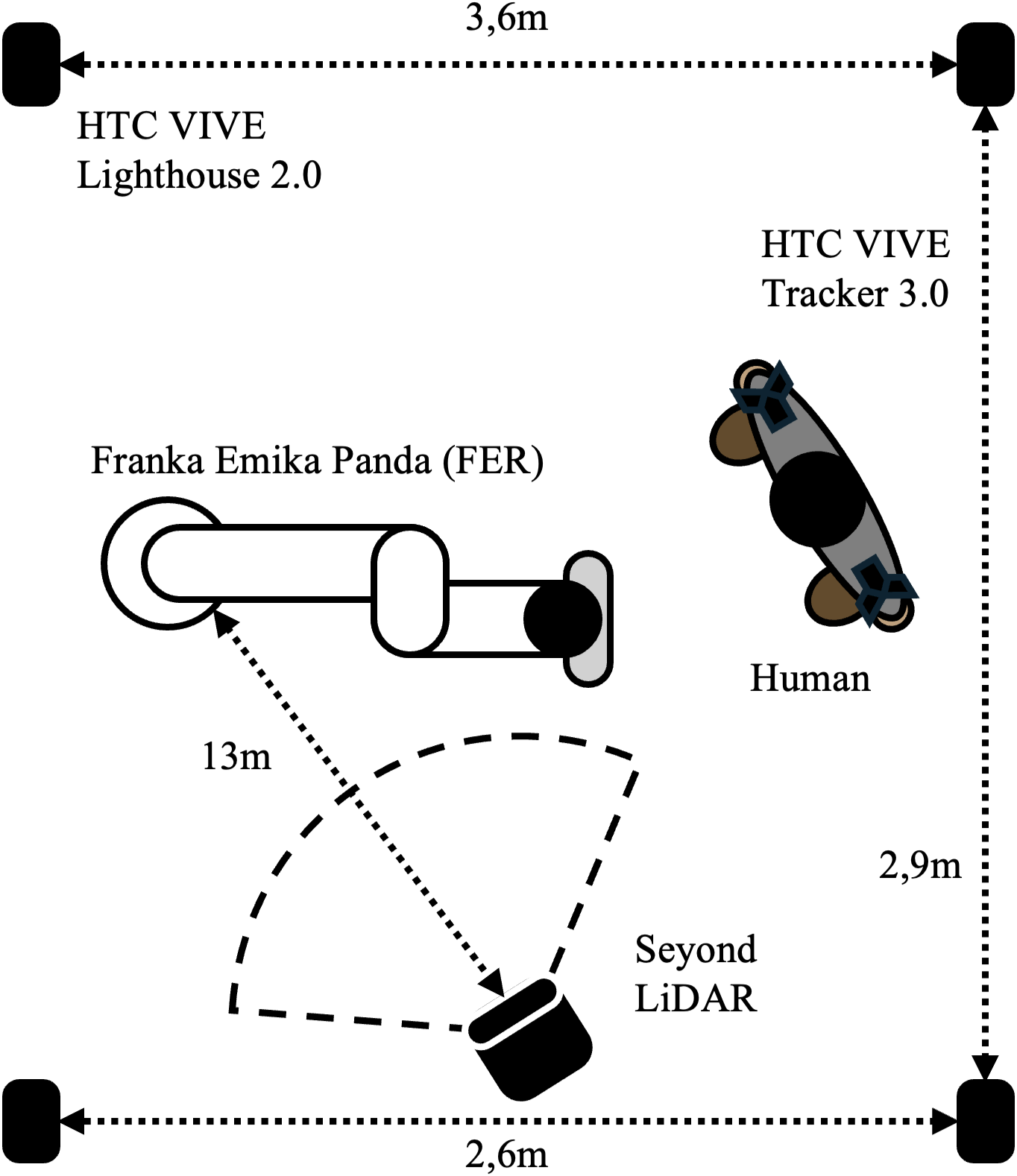} 
    }
    \caption{Schematic setup of the hardware to record LiHRA.}
    \label{fig:hardware-setup}
\end{figure}

Figure \ref{fig:hardware-setup} shows a schematic layout of the hardware setup we used to record the dataset in our laboratory. The four base stations are placed with a distance of roughly $3 m$ to each other and in a way that all trackers can reference themselves at any given time, because they are always in line of sight of at least two base stations. The LiDAR is placed at a height of approximately $1.9 m$ with a distance of $13 m$ to the robot, so it has the best possible view on the cobot and the human.

\subsection{Pipeline}
\label{sec:Pipeline}
For an accurate representation of the human keypoints and the joints of the robot in the point cloud, an extrinsic calibration of all three components, (i) LiDAR, (ii) cobot, (iii) MoCap system, to each other is necessary. Since the LiDAR has an intensity channel, reflectors can be used to accurately pinpoint a position in the real world to a position in the point cloud. An algorithm clusters all points with an intensity above a certain threshold and publishes its position with respect to the origin of the LiDAR coordinate system. Collecting at least three positions allows computing the orientation of an object using the Kabsch-Umeyama algorithm \cite{kabsch_solution_1976, umeyama_least-squares_1991}, given that the distance of the reflectors to each other is known. The Hand-Eye-Calibration method, calibrating the robot to the LiDAR, moves the robotic arm to different locations with a reflector attached. Through forward kinematics, the position of the reflector with respect to the robot's base is known, as well as the position with respect to the LiDAR through the tracking of the reflector. The robot is extrinsically calibrated using the rigid body transformation estimation method by Kabsch and Umeyama. With the calibrated robot and a custom mount, an HTC VIVE Tracker is attached to a known position on the robot to align the MoCap system with the LiDAR.

After calibration, the recording process takes place, which is accomplished using the Robot Operating System 2 (ROS 2) that also synchronizes all incoming point clouds, keypoints and joint states. The keypoints, joint states and point clouds are then exported to .json and .pcd files for easier integration in subsequent workflows.

\begin{figure}[thpb]
    \centering
    \framebox{
        \includegraphics[width=0.9\columnwidth]{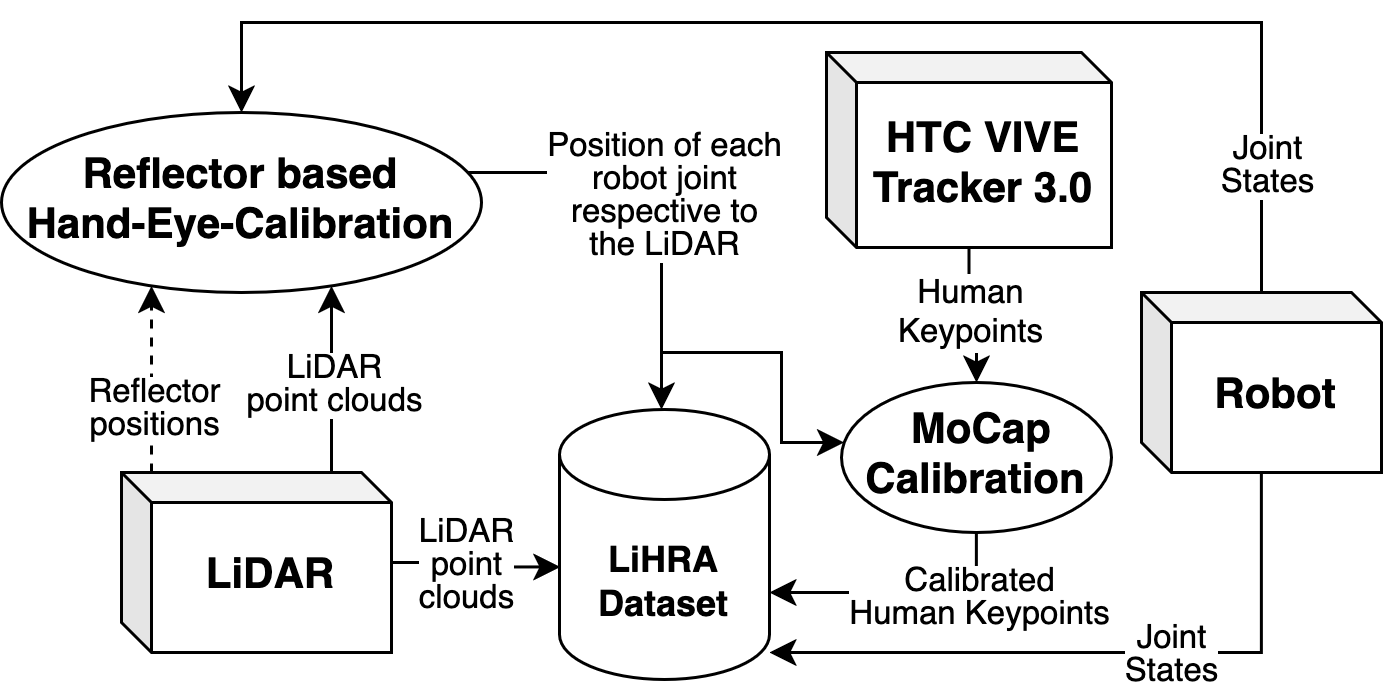} 
    }
    \caption{Soft- and hardware pipeline from the creation of the LiHRA dataset.}
    \label{fig:pipeline}
\end{figure}

\subsection{Scenarios}
\label{sec:Scenarios}
\begin{table*}[!ht]
\caption{LiHRA scenarios.}
\label{tab:scenarios}
\begin{center}
\begin{tabular}{c c c c c c}
\hline
\makecell{Scenario}                     & \makecell{Description}                                                & \makecell{Intentional\\Contact}   & \makecell{Unintentional\\Collision}   & \makecell{Frames} & \makecell{Framerate\\(Hz)}\\
\hline
\makecell{S1\\handover}                 & \makecell{Robot handing over three objects to the human.}                                       & \cmark                            & \xmark                                & 593               & 10.31\\
\hline
\makecell{S2\\dangerous\\handover}      & \makecell{Robot handing over two objects to the human,\\collision at the third handover.}               & \cmark                            & \cmark                                & 552               & 10.36\\
\hline
\makecell{S3\\collaboration}            & \makecell{Human polishing three objects presented by the robot.}                               & \cmark                            & \xmark                                & 1066               & 10.78\\
\hline
\makecell{S4\\dangerous\\collaboration} & \makecell{Human polishing two objects presented by the robot,\\collision at the third collaboration.}              & \cmark                            & \cmark                                & 841               & 10.77\\
\hline
\makecell{S5\\coexistence}              & \makecell{Robot moving, while human is inspecting from safe distance.}                    & \xmark                            & \xmark                                & 783               & 10.68\\
\hline
\makecell{S6\\dangerous\\coexistence}   & \makecell{Robot moving, while human is inattentive leading to collision.}   & \xmark                            & \cmark                                & 596               & 10.40\\
\hline
\end{tabular}
\end{center}
\end{table*}

After the calibration process, six different scenarios have been recorded (see Table \ref{tab:scenarios}). The scenarios involve in total three different actions with one scenario performing the action safely and if applicable only intentional contact (i.e., a contact of the human hand with the robot's gripper during a handover of an object) and one scenario performing the same action but with an unintentional collision. The actor in the dataset was instructed to act as realistic as possible, as if the robot hit the human completely accidentally.

In the first two scenarios, three small boxes ($20cm\times20cm\times6 cm$) are being picked by the robot and handed over to the human. The person then puts these boxes aside. The third handover results in a collision in the second scenario "S2 dangerous handover". During the collaboration, the person uses a cloth to polish the boxes from different sides, which are presented to the human by the robot. Because of the person being distracted, this leads again to a collision in the fourth scenario "S4 dangerous collaboration". The scenario of coexistence involves the person observing the robot's movements and then standing next to it, looking away but at a safe distance, out of reach in Scenario 5, but within reach in the last scenario, resulting in a collision.

\section{Experimental Evaluation}
\label{sec:Experimental-Evaluation}
An RM method is utilized to demonstrate the usability of the dataset. One of the modes of operation described in ISO/TS 15066:2016 is Power and Force Limiting (PFL), where the robot's velocity and applied forces are continuously monitored and controlled to ensure that any contact with a human remains within biomechanical safety limits \cite{araiza-illan_systematic_2016}. Drawing from this, we monitor the collision severity using the method described below in this section.

Additionally, we applied another continuous RM method proposed in \cite{katranis_2025}, which focuses on contextual data to retrieve hazard estimates based on velocities and distances. The results are to be found in the dataset, however, a detailed discussion of this method is beyond the scope of this paper.

\subsection{External Force Estimation}
To assess the severity of collisions in the given scenarios, we estimate the external forces acting on the robot's end-effector using classical model-based methods, which are well-established in the literature \cite{wu2017external, xu2024novel}. The external joint torques, $\tau_{\mathrm{ext}}$, serve as the basis for computing the external forces. The robot provides measured joint torque values, denoted as $\tau_\mathrm{meas}$, which are available in the dataset. By leveraging the robot's dynamic model, we compute the expected joint torques, $\tau_{\mathrm{model}}$. The total joint torque in the system consists of both internal and external contributions. The external torque can be obtained by calculating the residual between the measured and modeled torques:
\begin{equation}
\tau_{\mathrm{ext}} = \tau_{\mathrm{measured}} - \tau_{\mathrm{model}}.
\end{equation}
To obtain $\tau_{\mathrm{model}}$ we utilize the well-known dynamics equation:
\begin{equation}
\mathbf{M}(\mathbf{q}) \ddot{\mathbf{q}} + \mathbf{C}(\mathbf{q}, \dot{\mathbf{q}}) \dot{\mathbf{q}} + \mathbf{G}(\mathbf{q}) + \mathbf{F_f}(\dot{\mathbf{q}}) = \tau_{\mathrm{model}},
\end{equation}
where \( \mathbf{M}(\mathbf{q}) \in \mathbb{R}^{n \times n} \) is the mass/inertia matrix, \( \mathbf{C}(\mathbf{q}, \dot{\mathbf{q}}) \in \mathbb{R}^{n \times n} \) is the matrix representing the Coriolis and centrifugal forces, \( \mathbf{G}(\mathbf{q}) \in \mathbb{R}^{n} \) denotes the gravity torque, and \( \mathbf{F_f}(\dot{\mathbf{q}}) \in \mathbb{R}^{n} \) accounts for joint friction (e.g., Coulomb and viscous friction).  

The dynamic parameters of the robot have been computed and made available in \cite{gaz_franka_dynamic}. To compute $\tau_{\mathrm{model}}$, the joint accelerations, \( \ddot{\mathbf{q}} \), are needed.  Since direct measurements of joint accelerations are  unavailable, we estimate them by applying the Discrete Fourier Transform (DFT) on the available joint velocities. We utilize the differentiation property in the frequency domain, followed by an inverse DFT to recover the time-domain acceleration signal \cite{wu2017external}. 
In the final step, we calculate the Jacobian matrix and apply the pseudoinverse transformation, as the robot has 7-DOF, making the Jacobian matrix non-square. The external forces are given by:
\begin{equation}
\mathbf{F}_{\mathrm{ext}} = (\mathbf{J}^{\top})^{\dagger} \tau_{\mathrm{ext}},
\end{equation}
where \( (\mathbf{J}^{\top})^{\dagger} \in \mathbb{R}^{6 \times n}  \) is the Moore-Penrose pseudoinverse of the transposed Jacobian. This transformation provides a direct method for computing the external wrench acting on the end-effector from the estimated external joint torques.

\subsection{Results}
 \begin{figure}[thpb]
    \centering
    \framebox{
        \includegraphics[width=0.9\columnwidth]{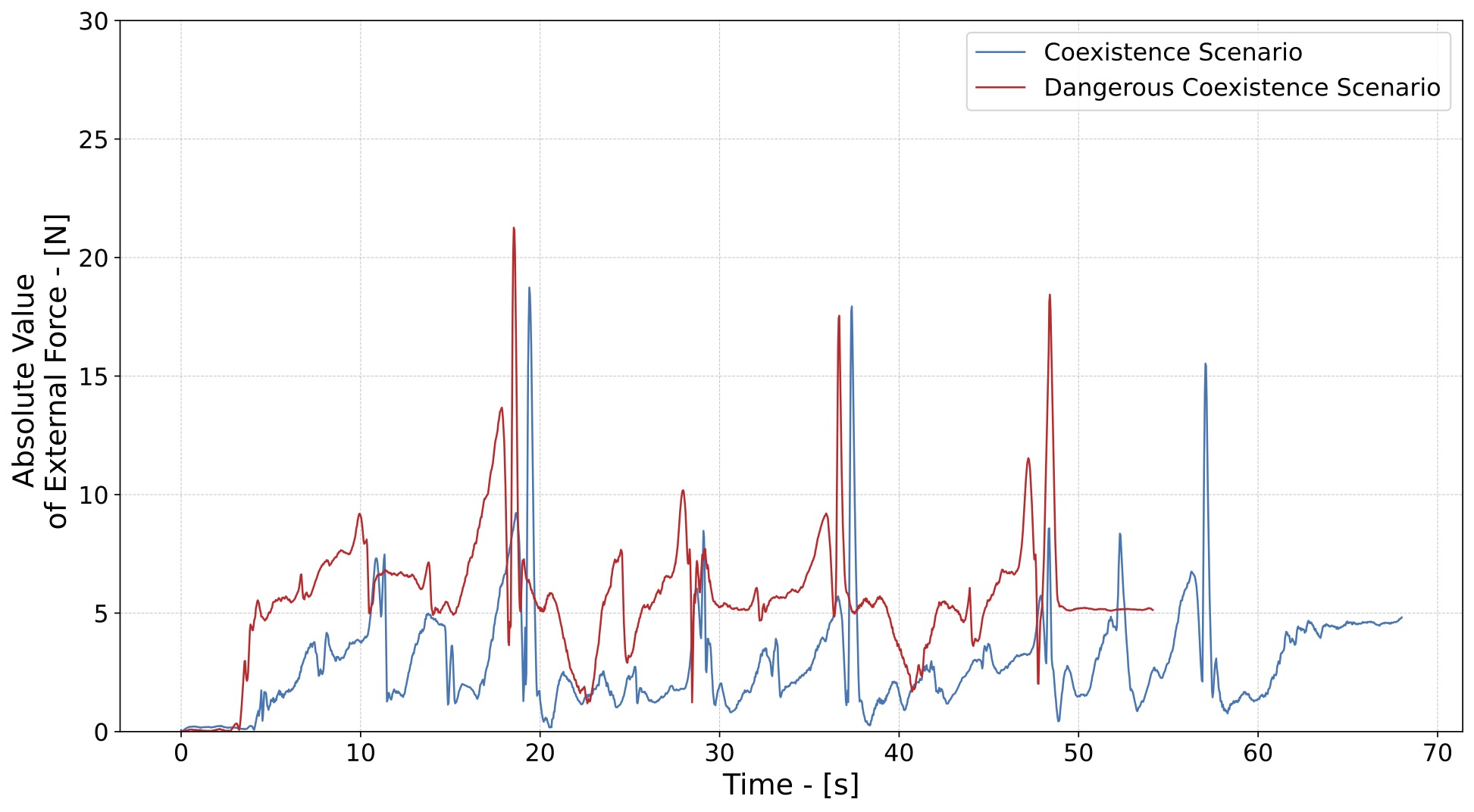}
    }
    \caption{Estimated external force magnitudes during the coexistence scenarios.}
    \label{fig:plot-coexistence}
\end{figure}

 \begin{figure}[thpb]
    \centering
    \framebox{
        \includegraphics[width=0.9\columnwidth]{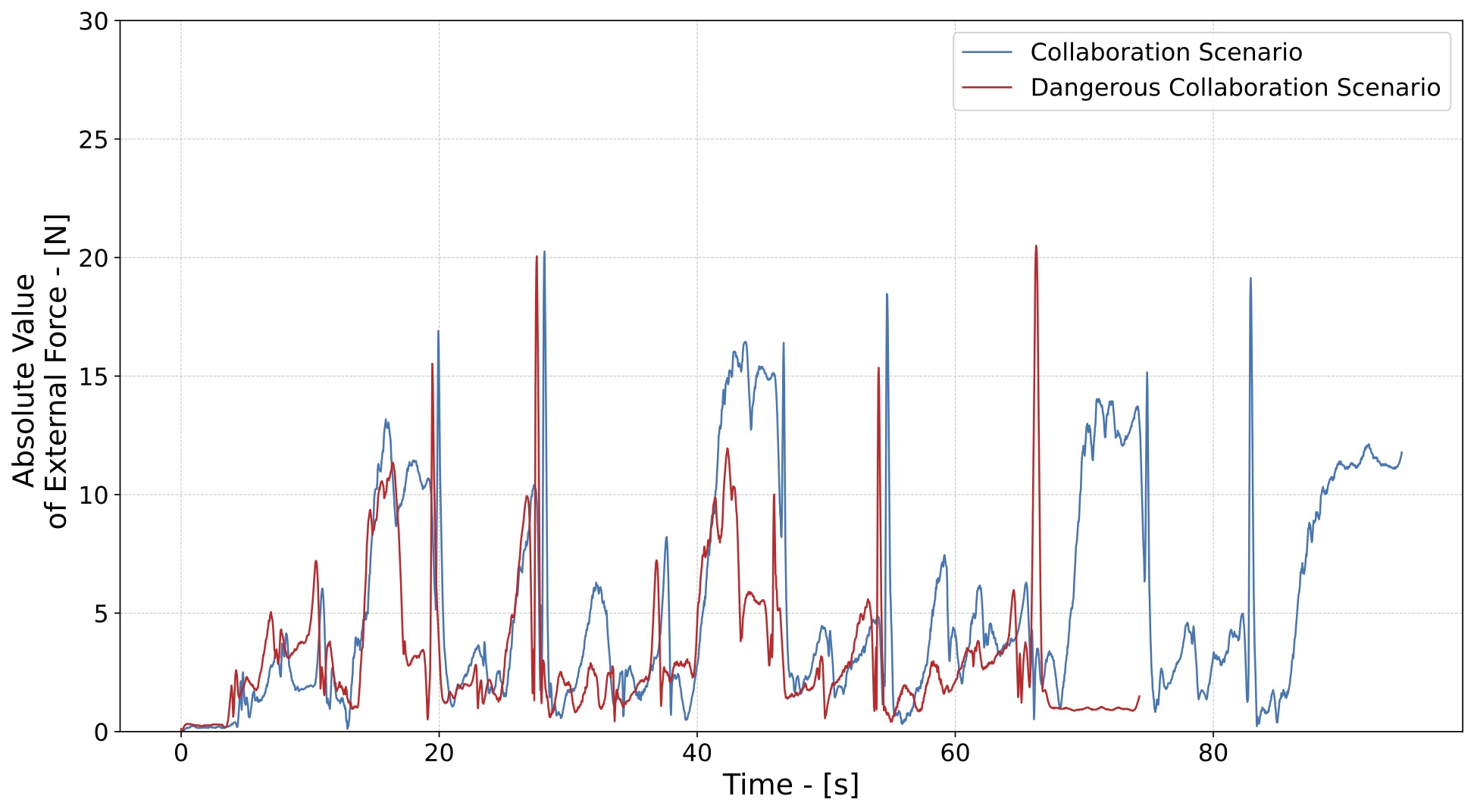}
    }
    \caption{Estimated external force magnitudes during the collaboration scenarios.}
    \label{fig:plot-collaboration}
\end{figure}


 \begin{figure}[thpb]
    \centering
    \framebox{
        \includegraphics[width=0.9\columnwidth]{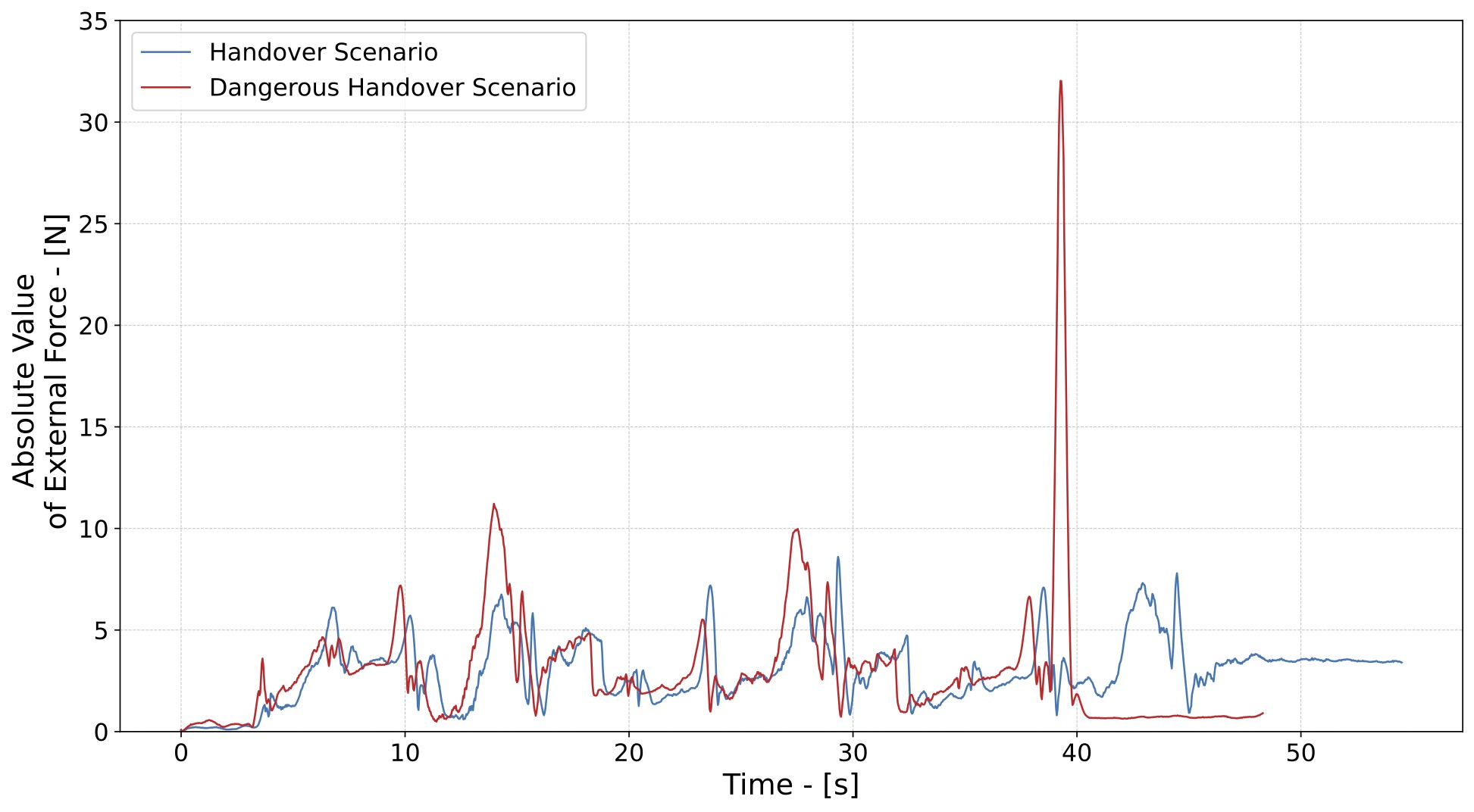}
    }
    \caption{Estimated external force magnitudes during the handover scenarios.}
    \label{fig:plot-handover}
\end{figure}



The external force magnitudes are compared to assess the impact of various interaction scenarios. As illustrated in Figures \ref{fig:plot-coexistence}, \ref{fig:plot-collaboration}, and \ref{fig:plot-handover}, the recorded forces exhibit characteristic peaks associated with different phases of interaction.

Several distinct force peaks are observed, which correspond to:
(1) intended physical contact during human-robot collaboration,
(2) the act of picking up the box and slight contact with the table surface,
(3) movements involving an additional payload (i.e., the picked-up box), and
(4) collisions occurring during the dangerous scenarios.

In particular, Figure \ref{fig:plot-handover} displays a pronounced peak corresponding to an unintentional collision, which significantly elevates the external force estimate. This peak is less evident in the other scenarios due to the compliant controller absorbing part of the impact and the robot executing a safety stop before high force levels were reached.

These results highlight the importance of context-aware risk and safety monitoring, which can differentiate between expected contact forces, such as those arising from intentional interactions, and potentially hazardous, unintended collisions.

A common characteristic across all scenarios is a constant offset of a few Newtons from the expected zero-force baseline. This offset arises from the gravitational force acting on the gripper, which intermittently carries a lightweight object. Since this factor is not explicitly modeled in the manipulator’s dynamic equations, it underscores the need for incorporating payload-aware modeling techniques. From a safety standpoint, unaccounted payload variations may lead to inaccurate force estimates, potentially masking or misclassifying external disturbances.

\section{Conclusions} 
\label{sec:Conclusions}
We presented LiHRA, the first dataset to enable learning based and advanced RA methods. LiHRA includes 3D LiDAR point clouds labeled with 5 human keypoints (head, hands and shoulders) captured by a MoCap system and robot joint states. The six recorded scenarios consist of three different HRI actions (coexistence, handover, collaboration) each executed once in a dangerous and non-dangerous manner. The key-contribution is the focus on the potential dangerous collisions. Additionally, a hazard identification method is presented to label not only the potential of a collision in each frame, but also the estimated force applied to the human. This enables the training of learning based methods to estimate these indices.

External force monitoring provides a reliable evaluation of collision severity. However, it is also important to accurately estimate the potential consequences of a collision between humans and robots. One possible approach is to assess the likelihood of injuries resulting from the applied forces. This subject is already being investigated by \cite{kirschner_towards_2024}.









\bibliographystyle{IEEEtran}  
\bibliography{bibliography}  
\end{document}